\documentclass[letterpaper, 10 pt, journal]{ieeeconf}  

\IEEEoverridecommandlockouts                              

\usepackage{graphics} 
\usepackage{epsfig} 
\usepackage{amsmath} 
\usepackage{amssymb}  

\usepackage{graphicx}
\usepackage[tight,footnotesize]{subfigure}
\usepackage{amsmath} 
\usepackage{amssymb}  
\usepackage{hyphenat}
\usepackage[table]{xcolor}
\usepackage{multirow}
\usepackage{url}
\usepackage{cite}
\usepackage{siunitx}

\def\floatsep{4.0pt}
\def\textfloatsep{0.0pt}
\def\dblfloatsep{4.0pt}
\def\dbltextfloatsep{4.0pt}

\newcommand\acomment[1]{\textcolor{red}{A:#1}}
\long\def\invis#1{}
\newcommand{\etal}{\textit{et al.}}

\newcommand\sect[1]{Section~\ref{#1}}
\newcommand\fig[1]{Figure~\ref{#1}}

\newcommand\eq[1]{Eq.~\eqref{#1}}

\title{\LARGE \bf
Underwater Surveying via Bearing only Cooperative Localization
}

\author{Hunter Damron, Alberto Quattrini Li,  and Ioannis Rekleitis
\thanks{The authors are with the Computer Science and Engineering Department, University of South Carolina, Columbia, SC, USA
        {\tt\small hdamron@email.sc.edu, \{albertoq, yiannisr\}@cse.sc.edu}}%
}

\begin{document}

\maketitle
\markboth{IEEE/RSJ International Conference on Robots and Intelligent Systems (IROS). Preprint Version. Accepted, 2018}
{}

\begin{abstract}
Bearing only cooperative localization has been used successfully on aerial and ground vehicles. In this paper we present an extension of the approach to the underwater domain. The focus is on adapting the technique to handle the challenging visibility conditions underwater. Furthermore, data from inertial, magnetic, and depth sensors are utilized to improve the robustness of the estimation. In addition to robotic applications, the presented technique can be used for cave mapping and for marine archeology surveying, both by human divers. Experimental results from different environments, including a fresh water, low visibility, lake in South Carolina; a cavern in Florida; and coral reefs in Barbados during the day and during the night, validate the robustness and the accuracy of the proposed approach. 
\end{abstract}

\section{INTRODUCTION}
The problem of {\em Cooperative Localization} (CL)~\cite{Rekleitis1998} has received a fair amount of attention in the robotics community over the years~\cite{Kurazume1994,burgard2000collaborative,Rekleitis2001c,Roumeliotis2002,Mourikis2006,leung_tro10}. It is described as the ability of a team of robots to utilize inter\hyp robot measurements in order to estimate the relative pose between vehicles and consequently constrain the pose uncertainty accumulation during operation. This is particularly important in  applications where there is neither access to a global positioning system, nor there is enough information in the environment to enable localization.  More formally, CL is concerned with the pose estimates of a team of two or more mobile robots which use sensory data for the purpose of enhanced localization accuracy compared to individual localization without cooperation. At the core of CL is the use of a sensor that provides information about the coordinate transformation matrix between two robots. CL has been used extensively for ground, aerial~\cite{RekleitisIROSWork2015}, surface~\cite{papadopoulos2010cooperative}, even underwater~\cite{bahr2009cooperative} robots. In this paper we focus on the underwater domain utilizing vision. 

The main motivation of this work derives from work on underwater cave mapping automation~\cite{RekleitisICRA2017b}. 
Cave mapping traditionally is performed by human divers who survey relative distances and orientations along segments of cave line that traverse the explored parts of a cave. The cave line represents a 1D ``roadmap'' inside the cave. However, the line is not located at the Voronoi diagram, also called the skeleton, or the medial axis~\cite{siddiqi2002hamilton}, of the cave, but where it was convenient for the cave explorers to attach the line to a fixed point. Central to this process is the estimation of the length and orientation of each segment between attachment points. The developed system can also assist in the underwater archeology domain for surveying submerged sites. The proposed approach consists of two devices used by two divers located at two points to record the distance and orientation between them; similar to the way two robots infer their relative pose.  Two underwater cooperative localization sensors have been constructed that can robustly produce relative pose measurements between them. These two devices can be mounted on underwater vehicles or deployed by divers. In this paper we describe the development of the two CL nodes, and the relative\hyp pose estimation algorithm as it pertains to the underwater domain. The proposed method is an extension of the 3D bearing only cooperative localization solution proposed by Dugas \etal~\cite{Dugas2013}. 

\begin{figure}
 \includegraphics[width=\columnwidth]{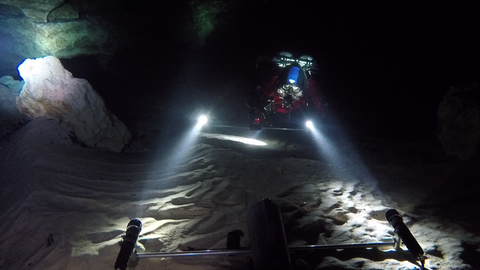}
 \caption{Underwater cooperative localization in a cavern in Ginnie Springs, FL, USA.\label{fig:uwcl}}
\end{figure}

More specifically, the proposed approach employs two cameras -- each equipped with two landmarks -- taking images of each other in a synchronized manner. The image ($I_A$) from camera {\em A}  contains the landmarks associated with camera {\em B}, and the  image from camera {\em B}  contains the landmarks associated with camera {\em A}. The two detected landmarks are registered as bearing measurements from each camera to the other system and an analytical geometry\hyp based solution provides the full 6{\em DoF} relative pose between the two cameras~\cite{Dugas2013}. The landmarks used in this work are dive LED lights that can provide adequate illumination to be detected in a variety of conditions.  In \fig{fig:uwcl}, Camera {\em A} is placed on the ground and camera  {\em B} is moved away; the experiment was conducted inside a cavern with no ambient illumination and the photo is taken by an outside observer. As can be seen in \fig{fig:uwcl}, underwater there are many challenges related to the image processing. In particular, in this image  the two landmark lights associated with camera B generate a light beam and there are reflections on the floor and on the diver. Additional data are used to assist in the outlier rejection process.  


In order to ensure the feasibility of the proposed approach to different applications, such as marine surveying and underwater cave mapping, the CL system was rigorously tested in a variety of environments ensuring great diversity of the lighting conditions. Experiments include a cavern zone, high turbidity fresh water, clear waters both during day and night time; for a detailed description please refer to \sect{setup}. 

The rest of this paper continues with a discussion of related work. \sect{sec:overview} provides a detailed overview of the proposed approach. Experimental results from different deployments underwater and in laboratory controlled conditions are presented in \sect{sec:results}. The paper concludes with a discussion of lessons learned and directions of future work.

\section{RELATED WORK}
\label{sec:bg}
The concept of CL was first introduced by Kurazume and Hirosi~\cite{Kurazume1994}, and the term was first used by Rekleitis \etal \cite{Rekleitis1998}. At a recent count, more than 100 papers have been published since then  covering many aspects. The CL approach has been used both in 2D and 3D, for localization~\cite{Kurazume1996}  and also for SLAM~\cite{Rekleitis2001c}. Using images~\cite{Dudek1996f,fox_collaborative}, LIDAR~\cite{howard2002,Rekleitis2003a}, and sonar~\cite{Grabowski2000}. The problem of CL with ``anonymous robots'' was presented by Franchi \etal \cite{Franchi2010}. More recently, Martinelli and Renzaglia~\cite{martinelli2017cooperative} provided the fundamental equations for fusing CL estimates with inertial data. 

Uncertainty propagation during CL was first given an analytical description in~\cite{RoumeliotisRekleitisAR2004}.
The initial formulation was based on the algorithm described in \cite{Roumeliotis2002} differing primarily in that robots had access to absolute orientation measurements instead of measuring their relative orientations; further studies of performance were presented by Mourikis and Roumeliotis~\cite{Mourikis2006}. 
Dieudonne \etal \cite{dieudonne2010deterministic} proved that for arbitrary relative measurements (range, bearing, and/or orientation) among robots, deciding if CL is possible is NP-hard. From a control perspective the observability~\cite{cristofaro20113d}, and consistency~\cite{Huang2009} of the problem was studied. More recently, Nerurkar \etal \cite{Nerurkar2009} and Leung \etal \cite{leung_tro10} proposed schemes of distributing the computations among a team of robots to improve computational efficiency of the algorithm. There is also a standard dataset available on\hyp line with different combinations of sensor measurements together with ground truth data~\cite{leung_ijrr11}.

Of particular interest is the analysis of different sensing modalities used in CL~\cite{Rekleitis2002c}. In particular, the effect of range~\cite{Trawny2009}, versus bearing measurements has been extensively analyzed~\cite{Zhou2010,Zhou2012}. In this work we focus on bearing only measurements as cameras are great protractors, providing better angle measurements compared to distance. Initially the problem was analytically solved in 2D~\cite{Rekleitis2012c}, and then the analytical solution was extended in 3D~\cite{Dugas2013}. At the same time, Dhima \etal \cite{dhimanIROS2013mutual} produced a numerical solution, clearly a more computationally expensive and less accurate formulation. The analytical solution was further used to assist the flying formation  of quadrotors~\cite{RekleitisIROSWork2015}.

This work utilizes the analytical solution to 3D analytical solution for cooperative localization in the underwater domain, extending the beacon detection method to better account for distortions underwater and incorporating additional sensor data for outlier filtering. Cooperative localization has been verified extensively above water. This paper provides experimental verification to the bearing only cooperative localization scheme in different underwater conditions. 

\section{SYSTEM OVERVIEW}
\label{sec:overview}
\begin{figure}[t]
 \includegraphics[width=\columnwidth]{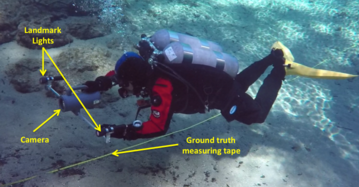}
 \caption{Underwater cooperative localization node with dive lights.\label{fig:node} }
\end{figure}

\begin{figure}[h]
 \fbox{\includegraphics[width=0.95\columnwidth]{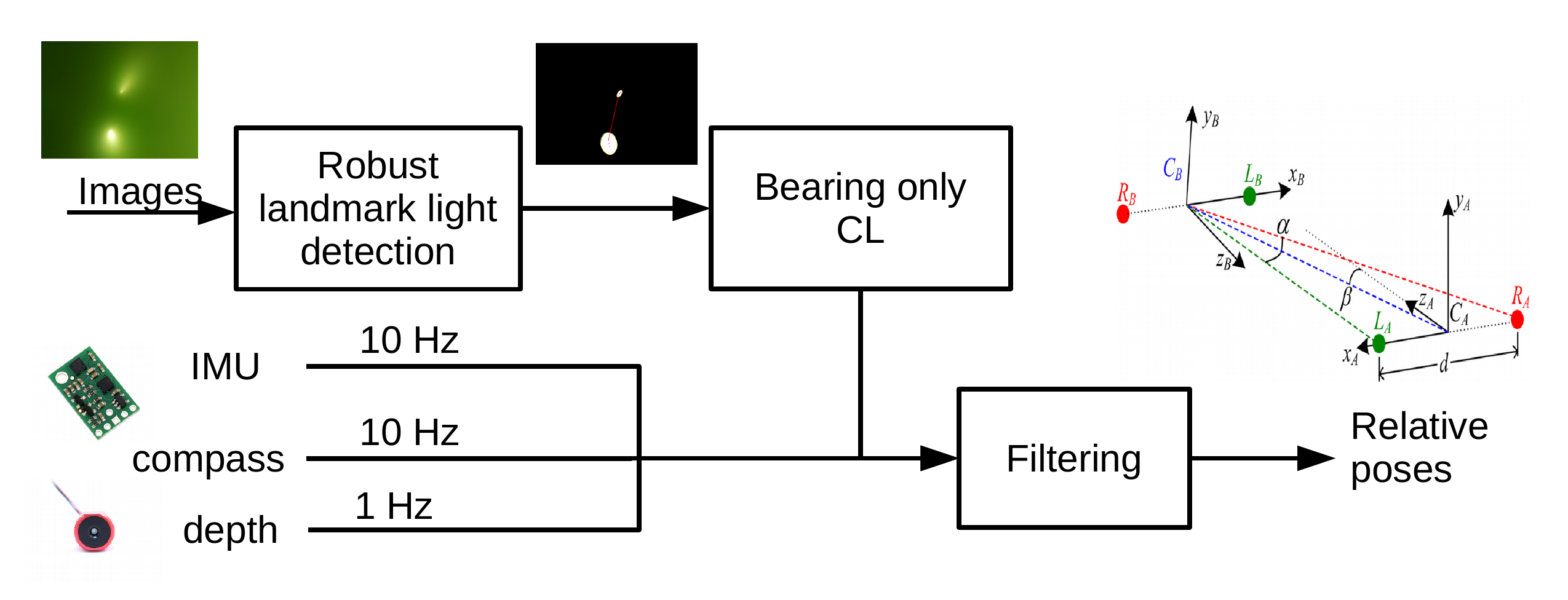}}
 \caption{Pipeline of the proposed approach.\label{fig:pipeline} }
\end{figure}

The proposed system consists of two sensors, termed nodes; {\em Node A} and {\em Node B}. Each node consists of a camera and two light landmarks; see \fig{fig:node} where one node in use is annotated. The two nodes have synchronized clocks and take images facing each other -- see \fig{fig:uwcl} -- at the same time. Concurrently, each node collects inertial, magnetic, and depth data from an IMU, magnetometer, and depth sensor; these data are used to further constrain the pose and attitude of each node. \fig{fig:pipeline} shows the pipeline of the proposed approach, starting from the input images, to the estimated relative pose. In the following, we describe each component in detail.

\begin{figure}[h]
\begin{center}
\leavevmode
\begin{tabular}{cc}
\subfigure[]{\includegraphics[height=0.1\textheight]{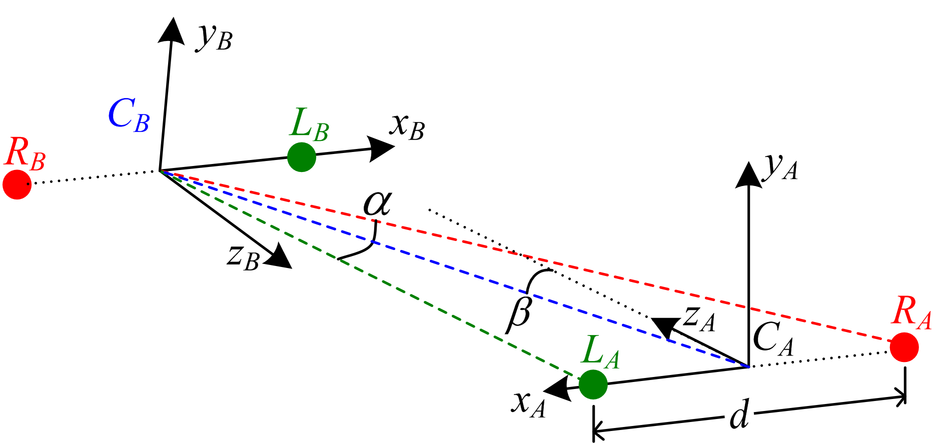}\label{fig:p2}}&
\subfigure[]{\includegraphics[height=0.1\textheight]{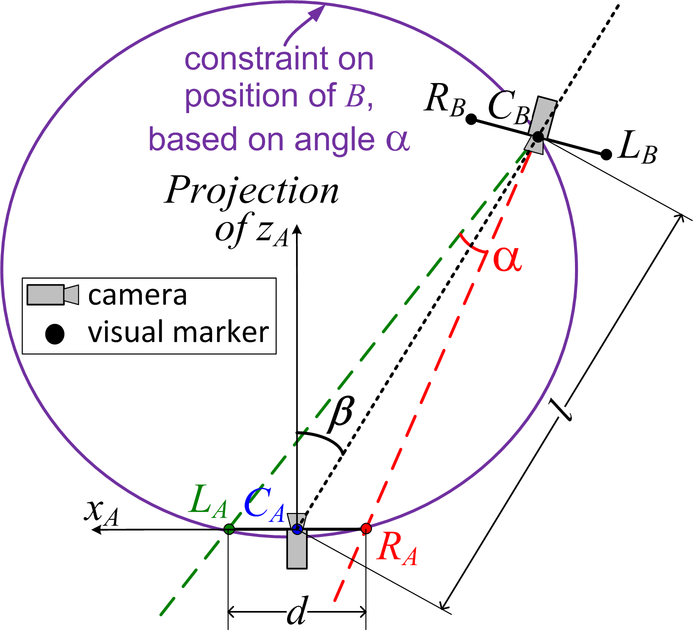}\label{fig:p3}}
\end{tabular}
\end{center}
\caption{Outline of the cooperative localization technique between two vehicles separated by an unknown distance $l$. The image taken by camera $B$ is used to measure the angle $\alpha$ between the two visual landmarks $L_A$ and $R_A$ distance $d$ apart, yielding a circular constraint. The relative angle $\beta$ is then measured by camera $A$. Camera $B$ is at the intersection of the circle and the line.}
\label{fig:schema}
\end{figure}

\subsection{Bearing only Cooperative Localization}
\label{sec:bearing}
For a complete treatment of the bearing only CL in 2D and 3D, please refer to the work of Giguere \etal \cite{Rekleitis2012c} and Dugas \etal \cite{Dugas2013}, respectively. For completeness sake, an outline of the approach will be presented next. The 2D case is based on the idea that the bearing measurement of the two landmarks, from camera A ($C_A$) constrains its position on a circle of fixed radius; the bearing measurement from camera B ($C_B$) constrains the position on a line; see \fig{fig:schema} for an illustration. For the 3D case, the main observation is that the collinearity of each camera and its landmarks produces a line; that line and a point, defined by the other camera, define a plane. Therefore, two cameras and two landmarks define a plane, and the 3D pose estimation can be performed utilizing the 2D constraints. More specifically, the relative pose between $C_A$ and $C_B$ can be analytically calculated by using two images ($I_A$ taken by $C_A$ and $I_B$ taken by $C_B$) recorded at the same time\footnote{The devices are synchronized at the beginning of the experiment (before submerging) by utilizing a Network Time Protocol (NTP) over an ad\hyp hoc Wi-Fi network, thus making it possible to extract images taken at the same time.}; see \fig{fig:schema} for the relationship between the coordinate systems of the two nodes. From these two images, the following data is obtained. First we extract two angles $\alpha$ and $\beta$:

\begin{figure*}[h!]
\begin{center}
\leavevmode
\begin{tabular}{cccc}
\subfigure[]{\includegraphics[height=0.125\textheight, angle =180]{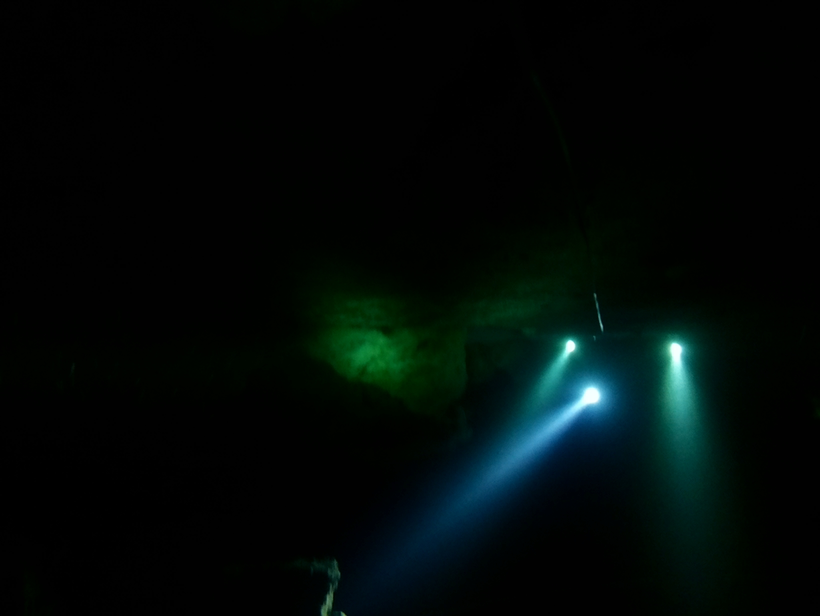}\label{fig:p1}}&
\subfigure[]{\includegraphics[height=0.125\textheight, angle =180]{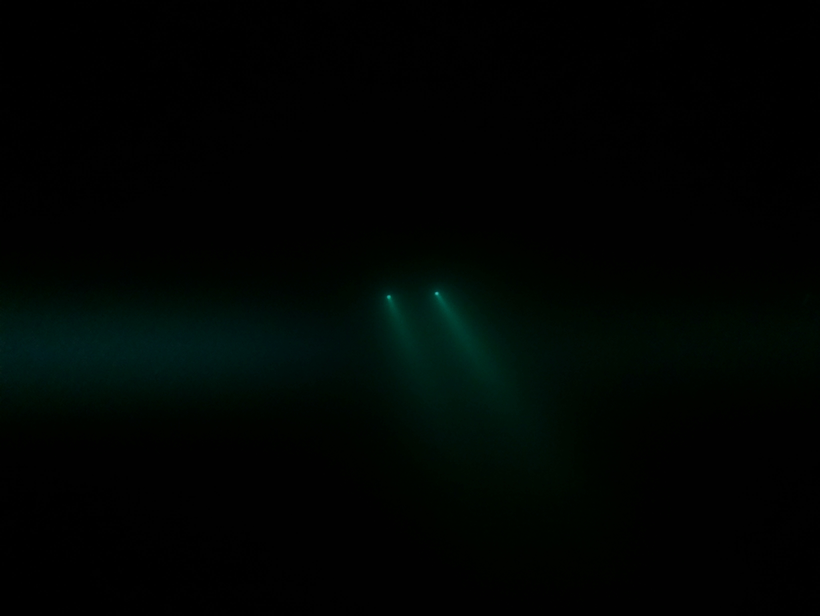}\label{fig:p2}}&
\subfigure[]{\includegraphics[height=0.125\textheight, angle =180]{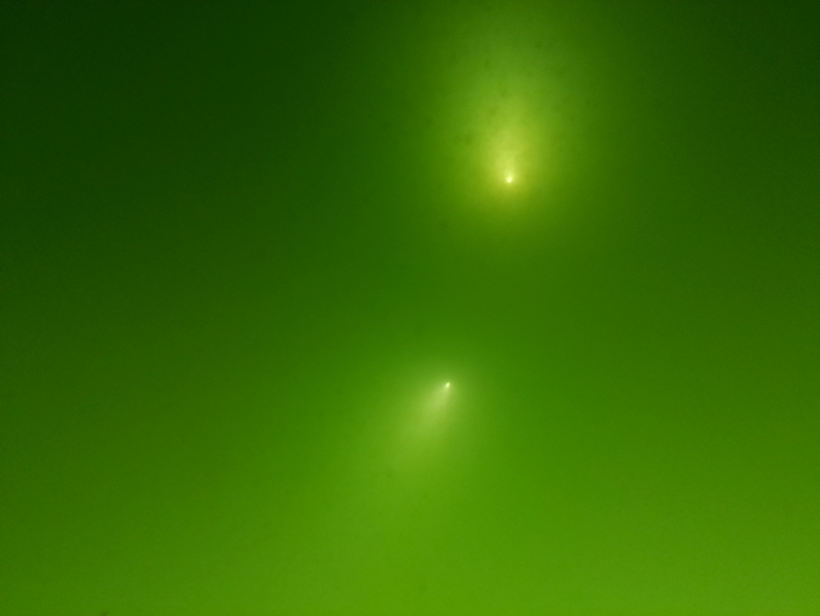}\label{fig:p3}}&
\subfigure[]{\includegraphics[height=0.125\textheight, angle =180]{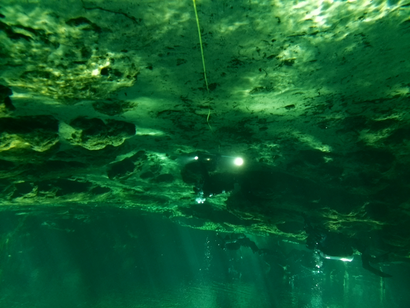}\label{fig:p4}}\\
\subfigure[]{\includegraphics[height=0.125\textheight, angle =180]{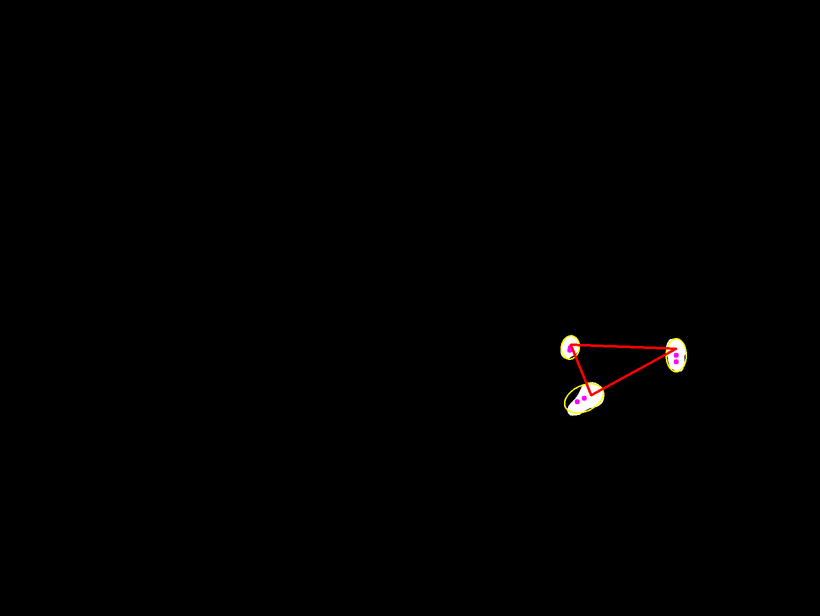}\label{fig:p5}}&
\subfigure[]{\includegraphics[height=0.125\textheight, angle =180]{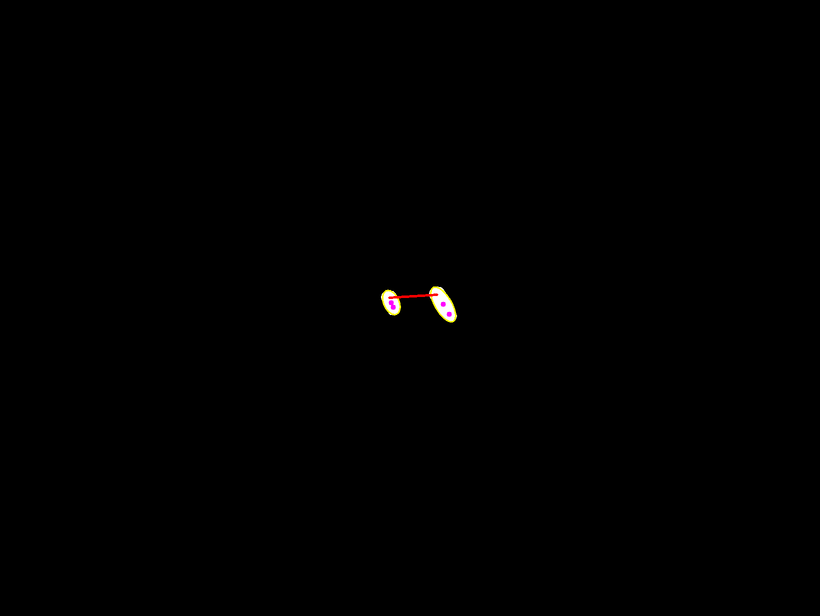}\label{fig:p6}}&
\subfigure[]{\includegraphics[height=0.125\textheight, angle =180]{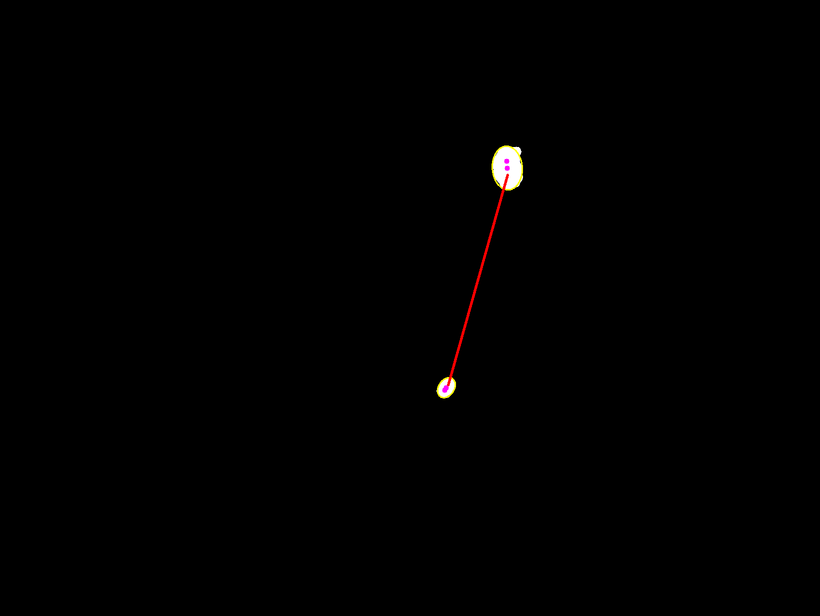}\label{fig:p7}}&
\subfigure[]{\includegraphics[height=0.125\textheight, angle =180]{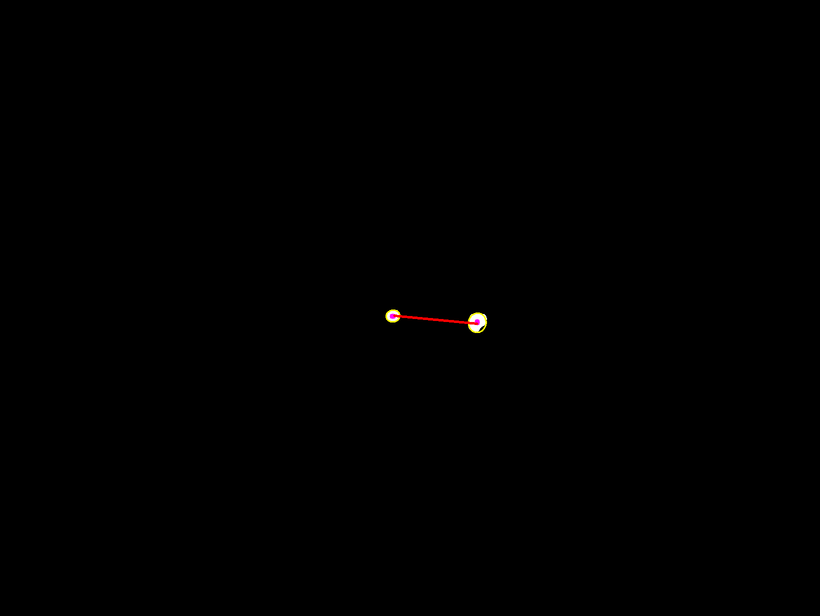}\label{fig:p8}}\\
\subfigure[]{\includegraphics[height=0.125\textheight, angle =180]{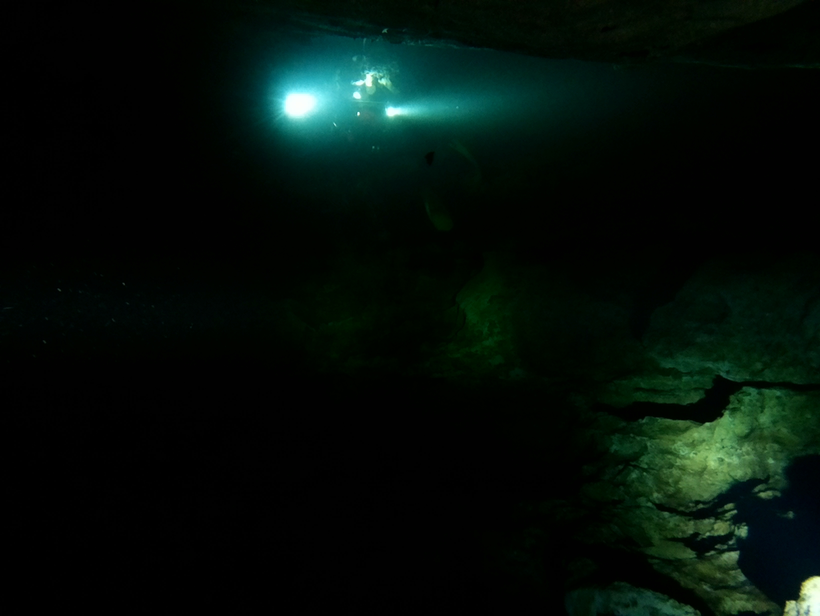}\label{fig:p9}}&
\subfigure[]{\includegraphics[height=0.125\textheight, angle =180]{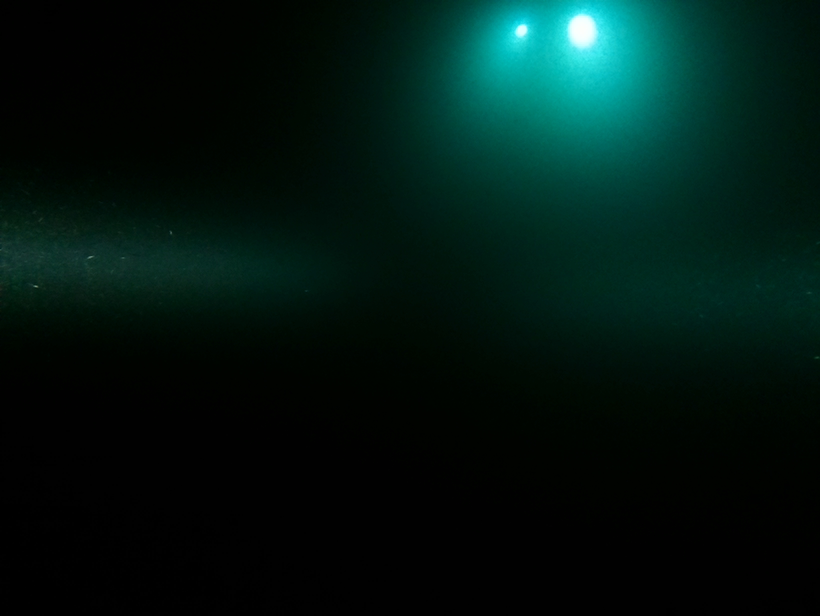}\label{fig:p10}}&
\subfigure[]{\includegraphics[height=0.125\textheight, angle =180]{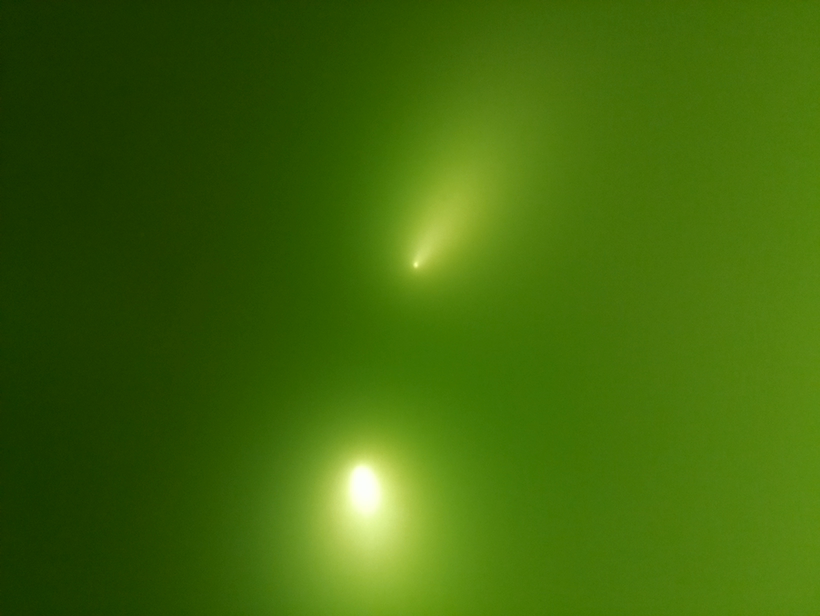}\label{fig:p11}}&
\subfigure[]{\includegraphics[height=0.125\textheight, angle =180]{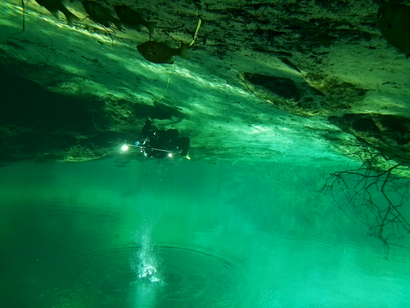}\label{fig:p12}}\\
\subfigure[]{\includegraphics[height=0.125\textheight, angle =180]{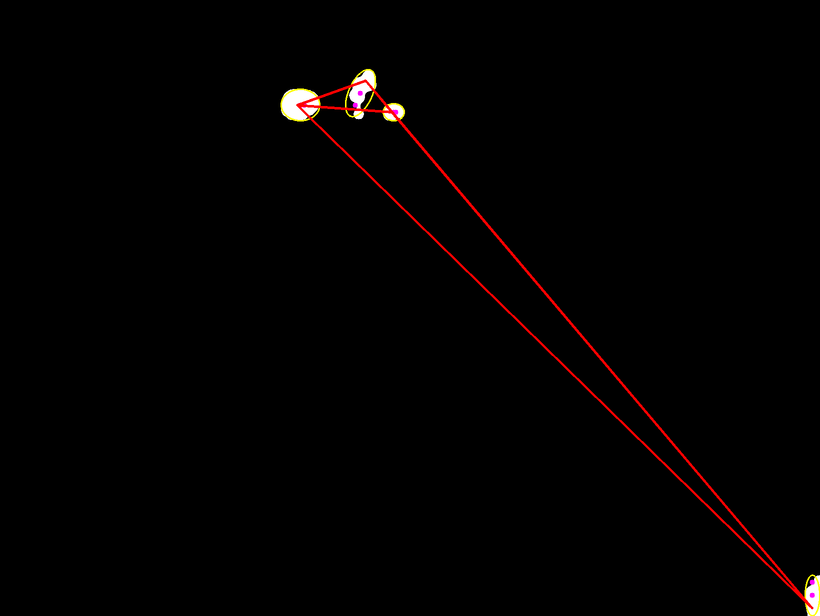}\label{fig:p13}}&
\subfigure[]{\includegraphics[height=0.125\textheight, angle =180]{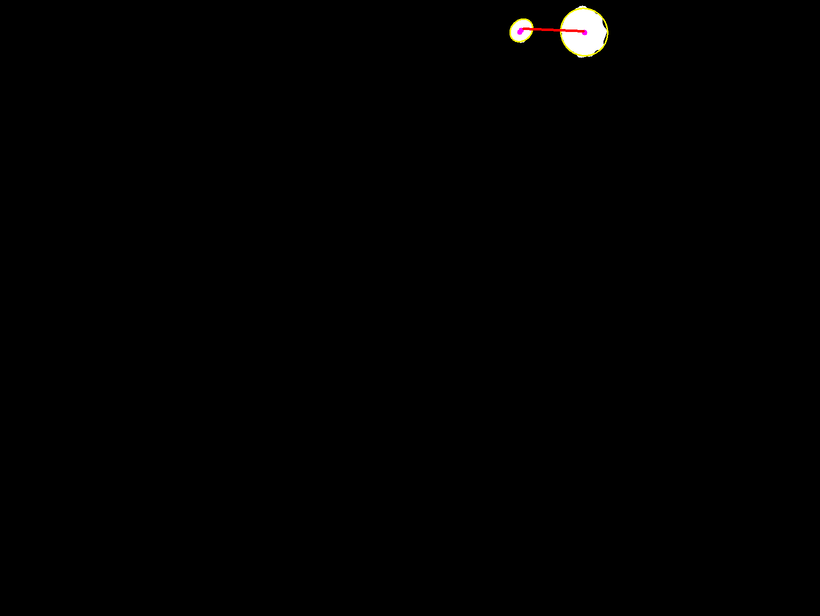}\label{fig:p14}}&
\subfigure[]{\includegraphics[height=0.125\textheight, angle =180]{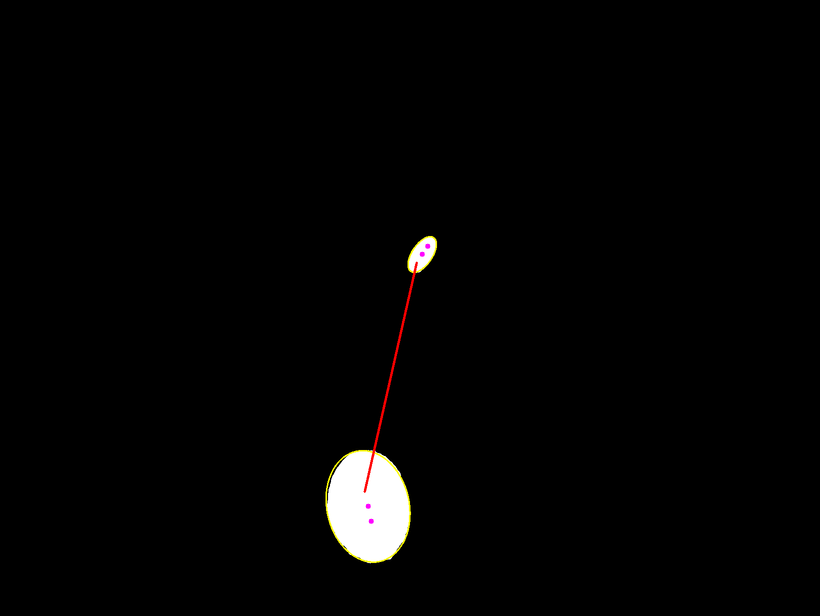}\label{fig:p15}}&
\subfigure[]{\includegraphics[height=0.125\textheight, angle =180]{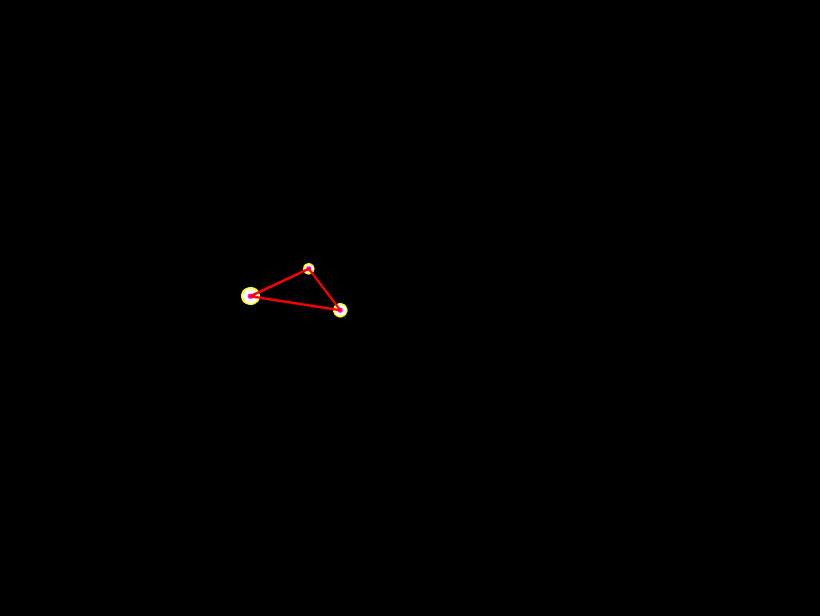}\label{fig:p16}}
\end{tabular}
\end{center}
\caption{Example images and results of blob detection system. In processed images, white regions are detected markers and red lines represent possible beacon pairs. {\bf Camera A}, raw images: (a) Ginnie Springs, Ballroom Cavern (Florida); (b) Night, Bellairs North reef (Barbados) (c) Lake Murray (SC); (d) Ginnie Springs, spring basin (Florida). {\bf Camera A}, blob (landmark) detection: (e) Ginnie Springs, Ballroom Cavern (Florida); (f) Night, Bellairs North reef (Barbados) (g) Lake Murray (SC); (h) Ginnie Springs, spring basin (Florida). {\bf Camera B}, raw images: (i) Ginnie Springs, Ballroom Cavern (Florida); (j) Night, Bellairs North reef (Barbados) (k) Lake Murray (SC); (l) Ginnie Springs, spring basin (Florida). {\bf Camera B}, blob (landmark) detection: (m) Ginnie Springs, Ballroom Cavern (Florida); (n) Night, Bellairs North reef (Barbados) (o) Lake Murray (SC); (p) Ginnie Springs, spring basin (Florida).}
\label{fig:env}
\end{figure*}

\begin{itemize}
\item from image $I_B$: $\alpha=\widehat{L_A C_B R_A}$, which is the angle between markers $L_A$ and $R_A$ about $C_B$;
\item from image $I_A$: $\beta$, the angle between the line passing through the origins of $C_A$ and $C_B$ relative and the optical axis of $C_A$, where the locations of ${L}_B$ and ${R}_B$ are used to approximate the position of $C_B$.
\end{itemize}

With these two angles $\alpha$ and $\beta$ and the known distance $d$ between markers on a robot, a closed\hyp form solution yields the distance $l=|C_AC_B|$ between the cameras~\cite{Rekleitis2012c}:
\begin{equation}
|C_AC_B|=l =\frac{d}{2 \sin\alpha} \left ({ \cos\alpha \cos\beta +
    \sqrt{1-\cos^2\alpha \sin^2\beta}} \right ).
\label{eql}
\end{equation}

An important fact pertaining to such an approach is that the majority of the uncertainty in the system will be on this distance $l$. This noisy distance estimate can be improved by performing the same computation described in \eq{eql} a second time, by extracting $\alpha$ from $I_A$ and  $\beta$ from $I_B$, and averaging the computed $l$'s.  The relative position [$x$,$y$,$z$] between cameras is then derived by extending the vector going from $C_A$ to the location of $C_B$ in the image frame to a length of exactly $l$. Sufficient information is contained in the two images $I_A$ and $I_B$ to recover uniquely the relative orientation between the two vehicles. It corresponds to a rotation matrix that: 

\begin{itemize}
\item aligns the perceived plane containing $C_B$, $L_A$ and $R_A$ with the perceived plane containing $C_A$, its right marker $R_A$ and the other camera $C_B$; and
\item aligns the perceived vectors $\overrightarrow{C_AC_B}$ in $I_A$ and   $\overrightarrow{C_BC_A}$ in $I_B$ in opposite directions.
\end{itemize}

\subsection{Underwater Vision for Accurate Blob Detection}
Vision processing underwater is much more challenging than in air due to light scattering from suspended plankton and other matter, which causes blurring and ``snow'' effects; loss of contrast; and loss of color information with depth. Moreover, the visibility conditions change with the time of the day, and the currents. The proposed approach was tested in different conditions as can be seen in \sect{sec:results} \fig{fig:env}(a-d). The influence of underwater conditions such as color loss~\cite{RekleitisBMVC2008}, blurring, and illumination changes has been studied by Oliver \etal \cite{Oliver2010}.

We propose a detection method which accounts for the distortions of light underwater by estimating the positions of markers from the visible cone of light they produce. Each image is converted to the HSV color space then thresholded. The threshold values are custom based on the environment as can be seen in \fig{fig:env}(a-d,i-l) where the lighting conditions are clearly different. The next step in each binary image is to identify the different blobs of light. First, morphological closing is applied and distinct regions are extracted from the binary image using contour detection. Then at the two ends of the bounding rectangle of each detected region the centroids of the brightest pixels are selected and compared to each other. The brightest side is assumed to be the one closest to the illuminating landmark. This is also apparent from observing not only the images presented in \fig{fig:env}(a-d,i-l) but also the external observer view in \fig{fig:uwcl}. In the case that a marker is not distorted significantly, both centroids are approximately equal to the center of the region. The above procedure results in a small number of landmark candidates. In particular during operation inside a cave environment or during the night where  there is no ambient light the divers carry additional lights, this results in additional blobs detected; see \fig{fig:env}(a), where there are three lights, and the corresponding \fig{fig:env}(e), where there are three blobs detected. Next the outlier rejection techniques are outlined which output the most plausible pair of landmarks for each image. The two pairs are then processed as described in the previous section and the relative pose is created.

\invis{
\begin{figure*}[th]
\begin{center}
\leavevmode
\begin{tabular}{ccc}
\subfigure[]{\includegraphics[height=0.12\textheight]{figures/NodeAnnotated}\label{fig:p2}}&
\subfigure[]{\includegraphics[height=0.12\textheight]{figures/NodeAnnotated}\label{fig:p3}}&
\subfigure[]{\includegraphics[height=0.12\textheight]{figures/NodeAnnotated}\label{fig:p1}}
\end{tabular}
\end{center}
\caption{Challenging conditions image from Camera A: (a) Ginnie Spring Cavern (Florida); (b) Lake Murray (South Carolina); (c) Bellairs North reef (Barbados). \acomment{tochange}}
\label{fig:tracks}
\end{figure*}
}

\subsection{Outlier Rejection}
A verification test, unique to our approach, is applied to all candidate pairs of markers in $I_A$ and $I_B$. As mentioned in \sect{sec:bearing}, there are two ways to calculate the distance $d$: either by using $\alpha$ from 2 candidate landmarks in $I_B$ and $\beta$ from the average (mid-point) of 2 marker candidates in $I_A$, or by doing the converse. The validity of a set of candidate markers is determined by the difference between the two estimates of $l$. Since $l$ in \eq{eql} is a closed-form solution, its computation time is low (less than \SI{300}{\ns} on a standard computer).

However, if many outliers are present in the images, additional data from magnetometer, IMU, and depth sensors are used to eliminate all outliers.
Contrary to most robotic applications where the presence of motors makes the magnetometer's measurements unreliable, in the proposed CL technique, the magnetic field is used to identify the azimuth of each node and to estimate the relative yaw between the nodes. In addition, the IMU is utilized to infer the roll and pitch of each device using measurements from the accelerometers. Lastly, depth sensor data provides an estimate of the relative depth. The collected measurements are then used to eliminate erroneous pairs of landmarks that appear as false positives in the previous processing.  See for example \fig{fig:env}(e) where three candidate markers were identified by the blob detection process, but the correct two markers were chosen by the outlier rejection system.

\section{EXPERIMENTAL RESULTS}
\label{sec:results}
Extensive experiments were conducted in different locations to ensure the robustness of the system.  In the following, first we present the hardware used and then describe the locations where the experiments were performed. Experimental results from different locations are described next and finally, we present a quantitative study conducted in our lab, using identical hardware while measuring the ground truth.

\subsection{Experimental Setup}
\label{setup}
\begin{figure}[h]
 \includegraphics[width=\columnwidth]{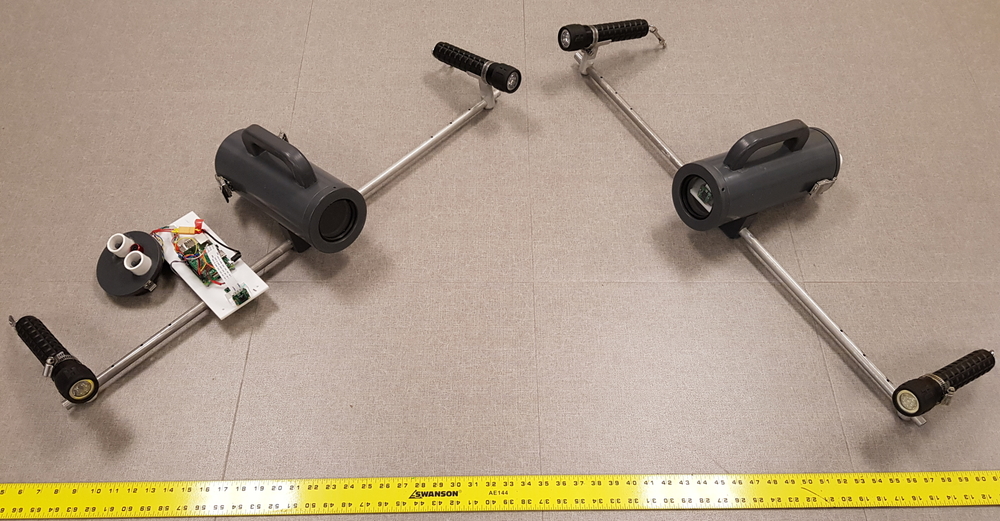}
 \caption{Underwater cooperative localization nodes with dive lights.\label{fig:system}}
\end{figure}

\paragraph{Hardware implementation} Two underwater nodes were constructed using a custom case waterproof case capable of reaching more than \SI{100}{\m} depth. The processing is based on a Raspberry Pi 3 computer connected to a Raspberry Pi Camera Module v2, a Pololu MinIMU-9 v3 IMU, and a Bar30 High-Resolution \SI{300}{\m} depth sensor. The design intentionally kept the cost low to ensure the adoption of the system by the underwater cave exploration and marine archeology communities. Two aluminum bars are rigidly attached, and two dive lights are attached on them; see \fig{fig:system} for the general appearance of the system. During experiments, the two landmark lights were kept at \SI{0.88}{\m} distance, however, they could be mounted in different position varying the spacing in between \SI{0.57}{\m} and \SI{0.88}{\m}.  

\paragraph{Test environments} In order to verify the versatility of the developed approach, the system was tested in a wide range of environments. In line with the primary application, the ballroom cavern at Ginnie Springs Florida was used to emulate a cave environment. No ambient light and clear water characterize this testbed; see \fig{fig:uwcl}. The nodes were tested at a depth of \SIrange{12}{15}{\m}. Similar conditions were encountered during tests at a night dive over the coral reefs of Barbados. The effect of ambient light was tested in three other scenarios. First, at the high turbidity waters of Lake Murray in South Carolina. The landmark lights produced long cones of illumination that needed to be addressed. The clear waters of Barbados' coral reefs and the spring fed waters of the basin outside the cavern at Ginnie Springs, FL, during the day produced a different set of challenges due to several false positives produced by the caustic patterns.

\subsection{Underwater Tests}
\begin{figure}[h]
\begin{center}
 \subfigure[]{\includegraphics[width=0.9\columnwidth]{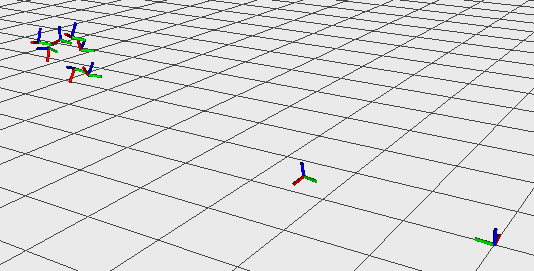}\label{fig:traj1}}\\
 \subfigure[]{\includegraphics[width=0.9\columnwidth]{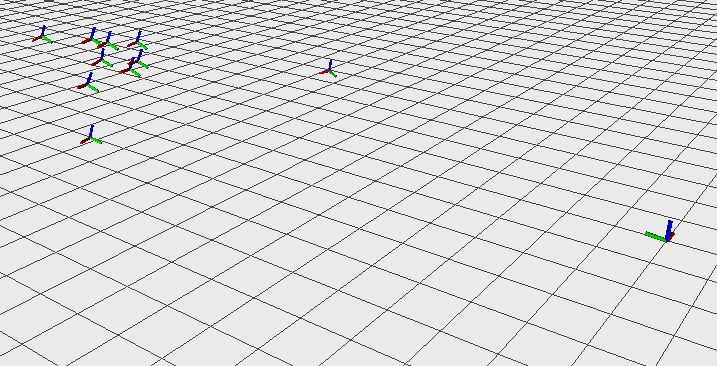}\label{fig:traj2}}\\
 \subfigure[]{\includegraphics[width=0.9\columnwidth]{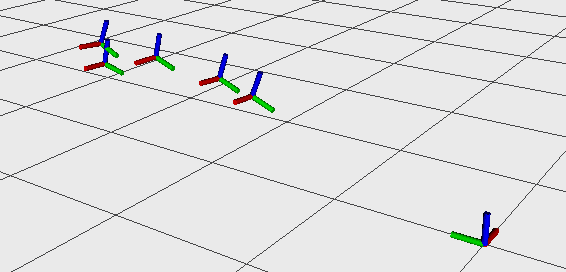}\label{fig:traj3}}\\
 \subfigure[]{\includegraphics[width=0.9\columnwidth]{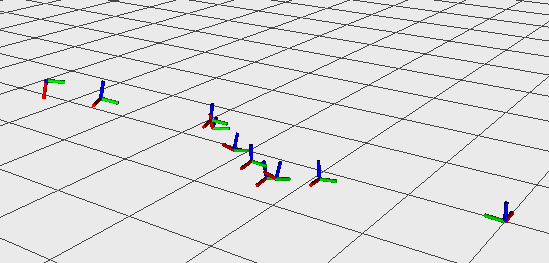}\label{fig:traj4}}
\end{center}
 \caption{Reconstructed trajectories, $C_A$ was kept still, $C_B$ was moved. (a) Ginnie Springs, Ballroom Cavern (Florida); (b) Day, Bellairs North reef (Barbados) (c) Lake Murray (SC); (d) Ginnie Springs, spring basin (Florida). Grid size 1 m.\label{fig:trajectories}}
\end{figure}

Different trajectories were tested in different environments. In all experiments, $C_A$ was kept stationary and $C_B$ was moved. The magnetic and inertial data are used to set the attitude of the stationary node. Note that, even though it was held to the ground, water movement had an effect especially on the nodes attitude. \fig{fig:trajectories}(a) presents a small segment of a trajectory inside the Ballroom cavern, where $C_B$ was moved back, and then moved in a circle to test different depths and orientations. The main challenge we observe with this dataset was the existence of additional light sources and sometimes reflection at the cave walls. \fig{fig:trajectories}(b) displays a trajectory collected during a night dive over a coral reef in Barbados. Due to the clear waters, the system was able to detecting the landmarks and produce the 3D relative pose, while $C_B$ was moved in different patterns. \fig{fig:trajectories}(c) has a short trajectory collected in Lake Murray, SC. While this was during a bright day, the visibility was really low due to particulates in the water, thus the landmarks disappeared after a short distance, even to the human eye. Finally, \fig{fig:trajectories}(d) displays a longer trajectory (approximately \SI{10}{\m}) collected at the basin fed from the clear waters of Ginnie Springs (just outside the cavern). The challenge here came from the caustic patterns at the bottom. 
However, as discussed earlier the outlier rejection ensures the correct pair of landmarks is selected. 

To verify the effectiveness of using sensors other than camera for additional outlier rejection, the number of correct marker selections was counted during both camera-only rejection and rejection using additional sensor data. During underwater tests, the camera-only rejection made \SI{72.4}{\percent} correct detections while the full system was \SI{84.46}{\percent} correct.  

\begin{figure*}[h]
\begin{center}
 \includegraphics[width=0.92\textwidth]{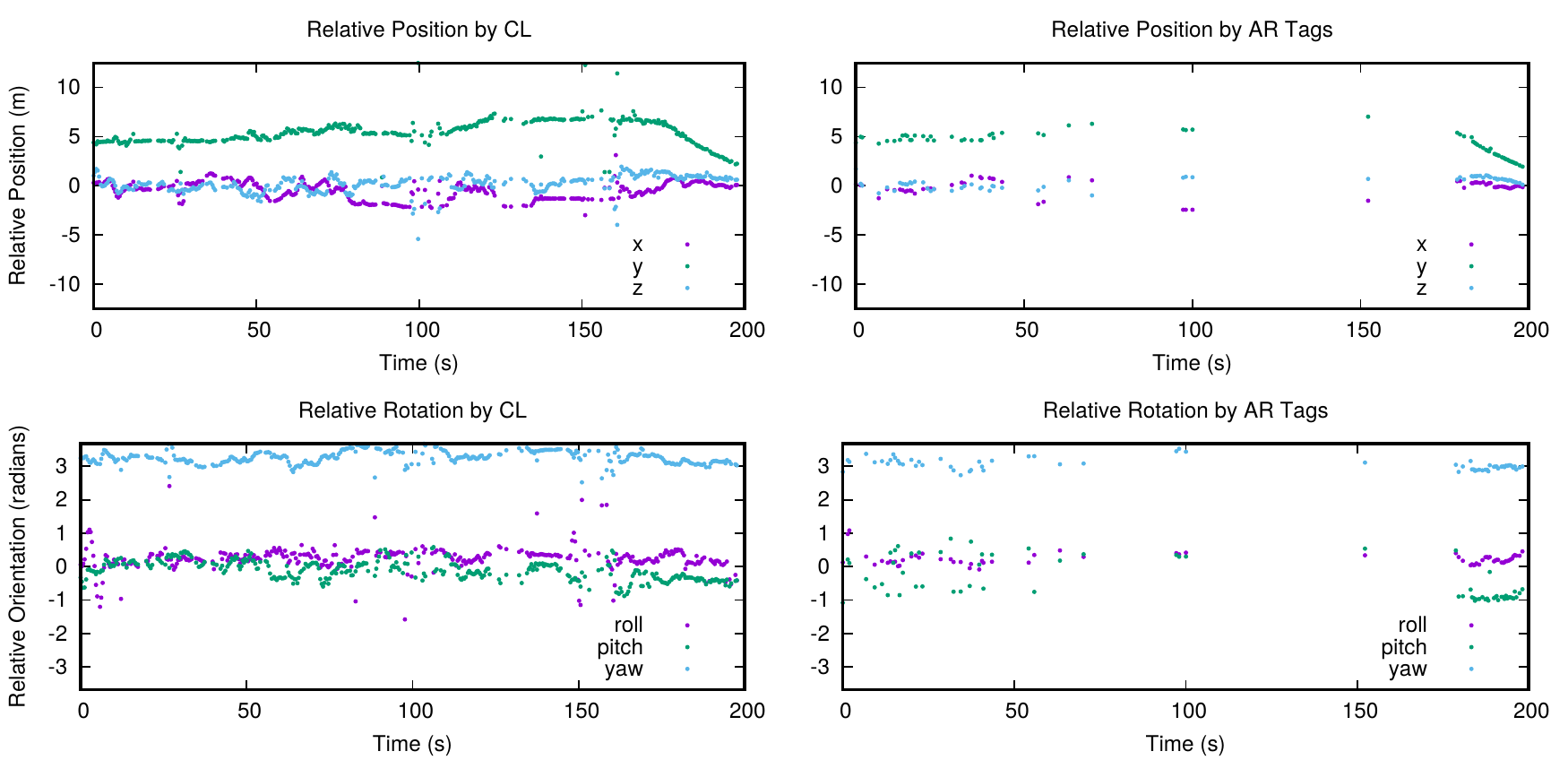}
\end{center}
 \caption{Relative pose calculated by both CL and by AR tag detection underwater. Both estimation schemes yield similar results with the exception of several outliers in CL detection when one beacon cannot be detected (\SI{88}{\s}, \SI{100}{\s}, \SI{160}{\s}). Note that AR tag detection was unable to provide a relative pose estimate for distances greater than about \SI{5}{\m}.}\label{fig:artagplots}
\end{figure*}

Because it is difficult to obtain an accurate ground truth pose estimate underwater, the CL system was compared to AR tag based cooperative localization~\cite{niekum2016artrack} for quantitative validation underwater. Two AR tags were attached to each node and used for relative pose calculation. The results of CL are compared to AR tag detection in \fig{fig:artagplots}. \invis{With the exception of false detections, the two methods generally differ by less than \SI{0.7}{\m} in relative position and \SI{0.5}{\radian} in rotation.} For tag size in the same scale as the sensors, the AR tags were less robust than the CL measurements with many missed estimates. Several outliers appear in the results of bearing only CL. This usually occurs when one beacon exits the camera's FoV in which case the outlier rejection does not have enough information to make the correct decision.

\subsection{Ground Truth above water}
\begin{figure}[h]
\begin{center}
 \includegraphics[height=0.23\textheight]{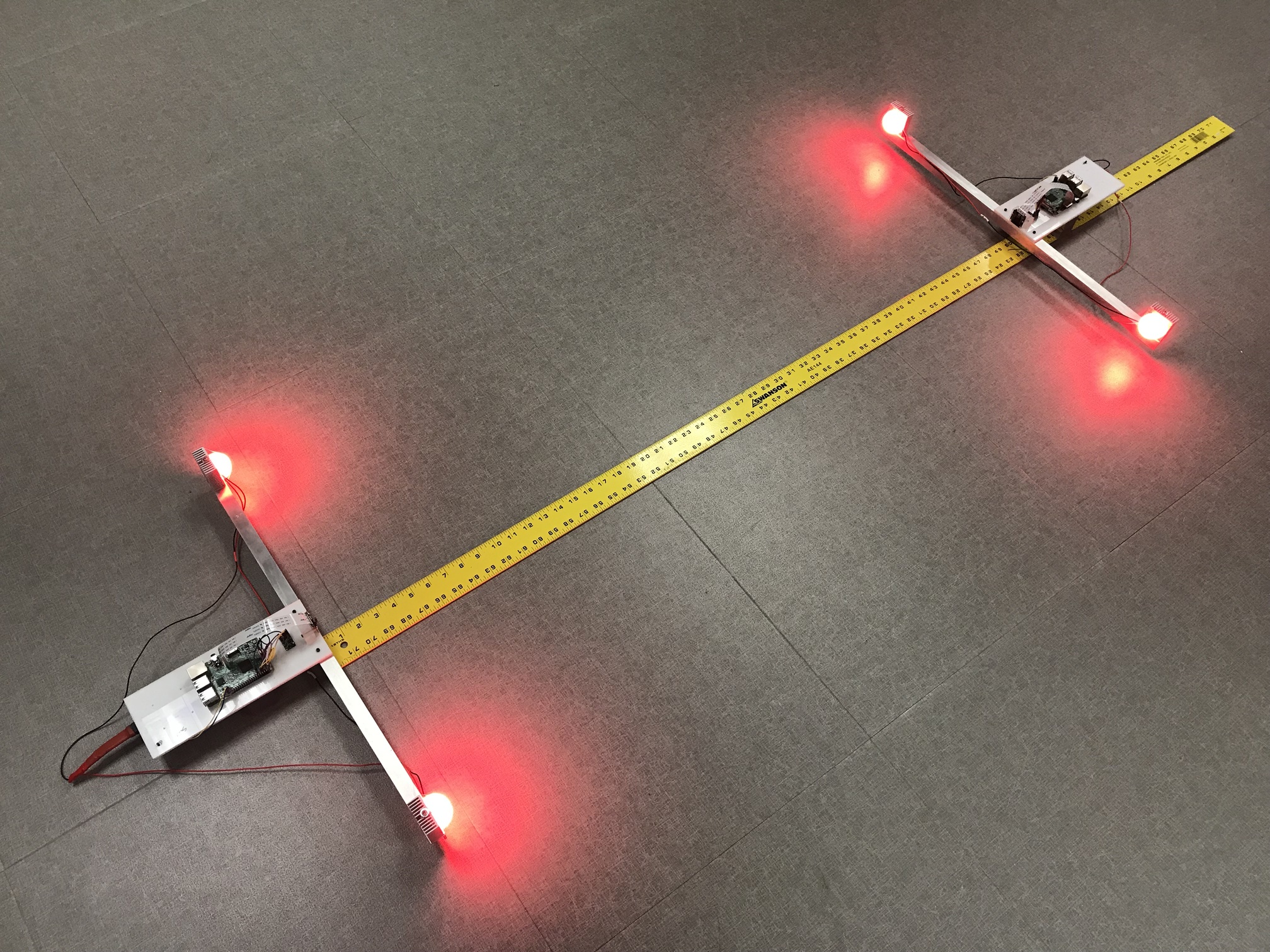}
\end{center}
 \caption{Hardware setup for test in our lab for ``ground truth'' validation.\label{fig:aboveCL}}
\end{figure}

\begin{figure}[h]
\begin{center}
 \includegraphics[width=\columnwidth]{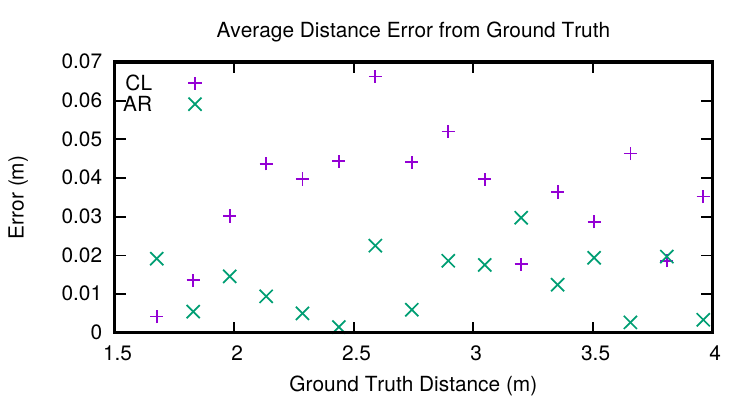}
\end{center}
 \caption{Average distance error from ground truth distance using both CL and AR tag relative pose calculation.\label{fig:GTError}}
\end{figure}

The identical hardware setup without the lights and depth sensor has been recreated for testing in the lab while establishing ground truth; see \fig{fig:aboveCL}. The different components have been tested separately, including the IMU parameters and the performance of the magnetometer. They were placed apart in fixed positions and the distance between them was measured using a measuring tape. AR tags were also used to calculate relative pose as validation. \fig{fig:GTError} presents a plot of the error between the calculated distance and the measured distance as a function of the measured distance between them for both CL and AR tag based estimation. The error is bounded within \SI{0.08}{\m} when the two nodes were moved from \SI{1.5}{\m} to \SI{4}{\m}. 

\section{CONCLUSIONS}
An analytical solution for 3D bearing only cooperative localization was augmented to operate underwater with the addition of IMU, magnetometer, and depth sensors. Challenging underwater conditions highlighted the effect of particulates in the water. As can be seen in most underwater images, the lights produced a beam with the brightest part at the source but with significant brightness all around. In addition reflections and the presence of other light sources produced several initial false positive blob detections; however, the outlier rejection introduced in this paper has been proven to be robust and ensures accurate pose estimates. 

For improved incorporation of IMU, magnetometer, and depth sensors in the future, the sensor data will be used in a multi-sensor fusion system rather than simply for filtering of candidate markers. This will allow a more fluid relative pose estimate and improve outlier rejection. 

We are currently discussing a collaboration with divers from the Woodville Karst Plain Project (WKPP)\footnote{\url{http://www.wkpp.org/}} for deploying the proposed system in the Turner Sink cave system in Florida. Future work will consider human factors for the deployment of this technology~\cite{RekleitisHFES2015}. The spacing between the landmark lights is crucial for achieving better accuracy over further distances. However, the turbidity of the water introduces additional constraints. We plan to analyze the relationship between landmark displacement and range to achieve the optimal arrangement for different visibility environments. Furthermore, we are considering deploying this system at a marine archeology dig in Greece to study its effectiveness. 

Deploying the proposed setup on different AUVs or on one AUV and a fixed point will enable the creation of a motion capture system underwater. Such an experimental setup will introduce much needed ground truth estimates for underwater applications.

\section*{ACKNOWLEDGMENT}
The authors would like to thank the National Science Foundation for its support (NSF 1513203, 1637876). We would like to thank fellow diver Lisa Jong\hyp Soon Goodlin for assisting in the Ginnie Springs dives and collecting some of the videos. 

\bibliographystyle{IEEEtran}
\bibliography{IEEEabrv,refs}

\end{document}